\documentclass{article}
\usepackage{spconf,amsmath,graphicx} 
\usepackage[
  top=0.75in,       
  bottom=1.05in,  
  left=0.65in,      
  right=0.65in      
]{geometry}

\usepackage[utf8]{inputenc} 
\usepackage[T1]{fontenc}    
\usepackage{hyperref}       
\usepackage{url}            
\usepackage{booktabs}       
\usepackage{amsfonts}       
\usepackage{nicefrac}       
\usepackage{microtype}      
\usepackage{xcolor}         
\usepackage{geometry} 
\usepackage{algorithm}
\usepackage{algorithmic}
\usepackage{subcaption}
\usepackage{float}
\usepackage{tabularx}
\usepackage{cleveref}
\usepackage{multirow}
\usepackage{appendix}

\usepackage{amsthm}
\graphicspath{{./figures/}{./}}
\def\etal{\emph{et al.}}

\newtheorem{theorem}{Theorem}
\newtheorem{lemma}{Lemma}

\newtheorem{definition}{Definition}

\title{Towards Trustworthy Federated Learning}
\name{Alina Basharat$^{\star}$ \qquad Yijun Bian$^{\dagger}$ \qquad Ping Xu$^{\star}$ \qquad Zhi Tian$^{\ddagger}$ }

\address{$^{\star}$  University of Texas Rio Grande Valley, $^{\dagger}$ University of Copenhagen, $^{\ddagger}$ George Mason University}

\begin{document}
%
\maketitle
\begin{abstract}
This paper develops a comprehensive framework to address three critical trustworthy challenges in federated learning (FL), i.e., robustness against Byzantine attacks, fairness, and privacy preservation. To improve the system’s defense against Byzantine attacks that send malicious information to bias the system's performance, we develop a Two-sided Norm Based Screening (TNBS) mechanism which allows the central server to crop the gradients that have $l$ lowest norms and $h$ highest norms. TNBS functions as a screening tool to filter out potential malicious participants whose gradients are far away from the honest ones. To promote egalitarian fairness, we adopt the $q$-fair federated learning ($q$-FFL). Furthermore, we adopt the differential private based scheme to prevent raw data at local clients from being inferred by curious parties. Convergence guarantees are provided for the proposed framework under different scenarios. Experimental results on real datasets demonstrate that the proposed framework is effective in improving robustness and fairness, while effectively managing the trade-off between privacy and accuracy. This work appears to be the first study that experimentally and theoretically addresses the fair, private, and robust issues in trustworthy FL.

\end{abstract}
\section{Introduction}
\label{sec:intro}

Federated learning (FL) has emerged as a promising paradigm to solve large-scale machine learning (ML) tasks, which allows multiple devices to collaboratively train a global ML model without sharing their private data and facilitates the collaborative improvement of a unified model through iterative updates from multiple nodes/clients \cite{10.1145/3298981}. Previous works on FL focus more on achieving high model accuracy or fast training convergence. Recently, the focus has been shifted more to emphasize the trustworthiness of FL to meet practical constraints \cite{jagatheesaperumal2024enabling}. 

To ensure trustworthiness in FL, several key obstacles need to be addressed. First, although the raw data remains on local devices in FL, the shared gradients or model parameters can inadvertently leak information about the underlying data by techniques such as model inversion attacks \cite{huang2021evaluating}. To address this issue, \cite{geyer2017differentially, wei2020federated, chen2024clfldp} achieve differential privacy (DP) by adding random Gaussian noise on the model update. Along this line, several works try to achieve exact convergence to an optimal solution to mitigate the negative impact of the added noise by adopting either the diminishing noise variance~\cite{huang2019dp} or utilizing a sequence of weakening factors to attenuate the added noise~\cite{wang2023tailoring}. Other approaches to preserve privacy include homomorphic encryption \cite{8859260,jin2023fedml} and secure multi-party computation~\cite {zhao2022pvd}.

Second, the iterative communication between local clients and the central server also exposes FL systems to threats such as Byzantine attacks \cite{xu2020towards} in which few malfunctioning nodes send altered or adversarial messages to the server \cite{10.1145/3335772.3335936}. Current defense mechanisms include the distance-based Krum and its variant Multi-Krum that select reliable updates based on the client's score, which is a summation of the pairwise Euclidean distance of its update and that of the remaining clients \cite{Blanchard}. The effects of median and trimmed mean techniques to mitigate the influence of attackers are studied in \cite{yin2018byzantine}, which validate the reliability of robust statistical methods in FL against adversaries. The norm based screening (NBS) method excludes suspicious attackers by cropping gradients with large norms \cite{zhou2024h}.

Lastly, standard FL approaches typically minimize an aggregate loss function without considering the uneven contributions and impacts of diverse data sources. This can inadvertently prioritize or disadvantage certain devices or groups, leading to biased models that perform well for some participants while failing others, resulting in unfairness \cite{li2019fair}. To address this, Mohri \etal~\cite{mohri2019agnostic} introduced Agnostic Federated Learning (AFL), which employs minimax optimization to reduce the training loss for the most disadvantaged client, ensuring that the model benefits all devices equitably. Zeng \etal~\cite{zeng2021improving} proposed reweighting the loss function in FedAvg to achieve a fairer classifier, showing improved accuracy over AFL. Fairness objectives differ by goal, with some aiming to balance disparities across larger and smaller groups, and others, like egalitarian fairness, seeking uniform performance across all users. Li \etal~\cite{li2019fair} and Zhou \etal~\cite{zhou2024h} introduced a fairness control parameter $q$ in the training objective to enhance the performance of poorly performing clients by amplifying their influence on global model updates.

This paper aims to develop a framework that simultaneously ensures fairness, robustness, and privacy. For client-level fairness, we adopt the $q$-fair FL scheme, which uses a fairness-promoting objective function to reweight the loss based on $q$ \cite{li2019fair}. To safeguard privacy, we apply DP by adding random Gaussian noise to model updates on the client side, preventing raw data inference. For robustness against Byzantine attacks, we introduce a two-sided norm-based screening (TNBS) method that filters out malicious participants by trimming gradients with the extremely lowest and highest norms. To the best of our knowledge, this is the first study to experimentally and theoretically address fairness, privacy, and robustness in trustworthy FL.\looseness=-1 

Specifically, our contributions include: 1) We recognize major trustworthy challenges in FL, i.e., fairness, privacy preservation, and robustness, and develop a comprehensive framework that simultaneously addresses these challenges to achieve fair, private and robust FL. 2) We establish certified fairness, privacy preservation, and robustness guarantees for the proposed method, even in the demanding non-iid settings, by confirming theoretical convergence guarantees under various assumptions regarding the model convexity. 3) We conduct extensive experiments on various real-world datasets and the experimental results validate the effectiveness of the proposed method in robustness, fairness, and privacy preservation.



\section{Proposed Method}
\label{Proposedwork}

\begin{figure}[htb]
  \centering
  \includegraphics[width=0.45\textwidth]{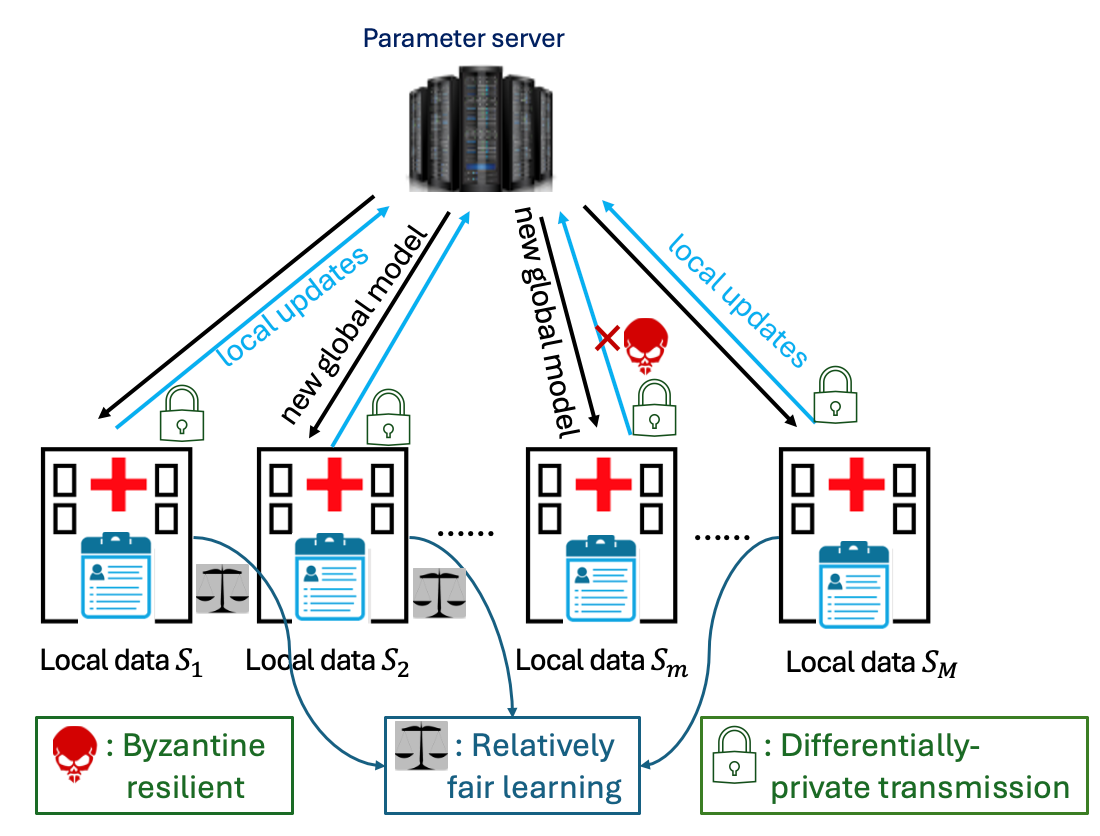}
  \caption{Trustworthy federated learning.}
  \label{fig:trust_fl}
\end{figure}

Consider the FL system is composed by $M$ clients and one central server, some of the clients are malicious attackers and the data distribution among all clients are non-iid. At a broader level, our algorithm is a version of distributed gradient descent distinguished by three key features: a fairness-promoting objective to ensure client-level fair model performance, a differentially-private message transmission scheme to prevent the disclosure of raw data, and two-sided norm-based screening to address Byzantine attack threats. The proposed framework is shown in Figure \ref{fig:trust_fl}.  Details of the three features are introduced below.

To achieve egalitarian fairness, \cite{li2019fair} incorporates a parameter $q$ to reweight the original loss function $F_i(\theta)$ parameterized by a common $\theta$ from local clients. The process starts with an initialized model $\theta_0$ and continues for a specified number of iterations $T$. At each iteration $t$, the server sends the current model parameters $\theta_t$ to all participating clients. Each client $i$ receives the model parameters $\theta_t$ and collaboratively optimizes a local fairness-aware objective function $H(\theta_t)$:
\begin{equation}
\label{fairnesseq}
H(\theta) = \sum_{i=1}^{M} \frac{p_i}{q+1} F_i^{q+1}(\theta), 
\end{equation}
where $p_i$ is the original weight for client $i$, which is typically set to be $p_i = \frac{n_i}{n}$ with $n_i$ being the number of samples client $i$ holds and $n$ being the total number of samples across all clients. In standard FL, the weight $p_i$ ensures that clients with more data have proportionally greater influence on the training process, which however, results in unfair training results when the data distribution is unbalanced across clients. Therefore, a tunable parameter $q$ is introduced to adjust the emphasis on fairness by modifying how losses from different clients are reweighted. A higher $q$ value increases the emphasis on clients with higher losses, aiming for a more uniform accuracy distribution across devices.

Each client $i$ then has a new loss function $H_i(\theta) = F_i^{q+1}(\theta)$ and uses its local data to compute the gradient:
\begin{equation}
\label{gradient}
g_i = \nabla H_i(\theta),
\end{equation}
which is then be transmitted to the central server for aggregation.

However, the transmission of gradients are still likely to leak the underlying raw data by techniques such as model inversion attacks \cite{huang2021evaluating}. Thus, we propose to utilize differentially-private (DP) encryption for message transmissions. DP encryption in FL acts as a safeguard that guarantees the model does not provide any information about the data of any client. Mathematically, DP can be defined as follows.
\begin{definition}
A randomized mechanism \( \mathcal{M}: D \to R \) satisfies \((\epsilon, \delta)\)-DP if, for any two adjacent datasets \( D, D' \in D \) that differ by a single data point (i.e., the data of one individual), and for all subsets of outputs \( S \subseteq R \), it holds that:
\begin{equation}
\label{DP}
 \Pr[M(D) \in S] \leq e^\epsilon \Pr[M(D') \in S] + \delta, 
\end{equation}
where \(\epsilon\) represents the privacy loss parameter, a smaller values denote stronger privacy;  \(\delta\) allows for a small probability of error and is typically set close to zero.
\end{definition}

After obtaining the updates from \eqref{gradient} at client side, each client encrypts its gradient with a random Gaussian noise vector $n\sim \mathcal{N}(0, \sigma^2)$:  
\begin{equation}
\label{DP2}
     \tilde{g}_i = g_i + n,
\end{equation}
where the variance is controlled by a sensitivity parameter $C$ that ensures \((\epsilon, \delta)\)-DP as follows,
\begin{equation}
     \sigma^2 = \frac{2C^2 \ln(1.25/\delta)}{\epsilon^2}   \,.
\end{equation}

Once the users calculate local gradients, \cite{zhou2024h} suggest employing NBS to combine all these gradients before utilizing the result to modify the model that excludes the gradients with significant norms and averaging the rest to produce the output. However, NBS assumes gradients with norms below threshold is honest, which might not be true. Some nodes can send arbitrarily small gradients to bias the learning process, which motivates the development of the two-sided NBS. TNBS retains only those gradients whose norms fall within the specified quantiles $p$. 

Denote the gradients received from clients as $g_1, g_2, \ldots, g_M$, TNBS first computes the norms of all clients' gradients and sorting them in an ascending order, resulting in a new sequence where $\|g_{(1)}\| \leq \|g_{(2)}\| \leq \ldots \leq \|g_{(M)}\|$. Then the upper and lower bounds for screening based on the quantile parameter $p$ that determines how many gradients will be kept are obtained: the upper quantile $U_{\text{p}} = 1 - \frac{p}{2}$ and the lower quantile: $L_{\text{p}} = \frac{p}{2}$.
%
 



Using these quantiles, TNBS computes the upper and lower thresholds by multiplying the quantile values by the number of clients $M$, giving the threshold indices: $Q_{\text{high}} = U_{\text{p}} \times M $ and $Q_{\text{low}} = L_{\text{p}} \times M.$ Then, TNBS crops the Byzantine attackers by only keeping those gradients whose index fall within the computed thresholds, i.e.,  $Q_{\text{low}} \leq (i) \leq Q_{\text{high}}$.

The remaining gradients are averaged into an aggregated gradient $G$ for gradient descent:
\begin{equation}
\textstyle \theta_{t+1} = \theta_t - \eta \times G ,
\end{equation}
where $\eta$ represents the learning rate.
%

\section{Theoretical Analysis}

In this section, we first establish a fundamental theoretical property of the proposed method. We then formulate two principal convergence theorems.

\subsection{Preliminaries}
In section \ref{Proposedwork}, we discussed how TNBS intuitively mitigates the influence of Byzantine participants in federated settings.
To further substantiate the effectiveness of proposed mechanism under the conditions of Byzantine attacks, fair aggregation and the integration of DP, we introduce Lemma \ref{thm:lemma,1} to provide a mathematical validation of the resilience and fairness metrics.


\begin{lemma}
\label{thm:lemma,1}
Suppose that a percentage of the local gradients, denoted by $\alpha$, are Byzantine and the index set of honest gradients is denoted as $\mathcal{M}$. Additionally, DP is implemented by adding Gaussian noise with variance $\sigma^2$ to each honest gradient. Let $G = \text{TNBS}_p$ be the aggregated gradient after applying the TNBS, then the following inequality holds \cite{zhou2024h} :
\begin{equation}
\label{lemma1}
\|G - \nabla H\| \leq \frac{2\alpha}{1-p} \|\nabla H\| + \max_{i \in M} \|g_i - \nabla H\| \, + \sigma ,
\end{equation}
where $\nabla H := \frac{1}{M} \sum_{i=1}^{M} \nabla H _i$ is the aggregated gradient and  \( \nabla H_i \) is the honest gradient of user \( i \).
\end{lemma}



\subsection{Main theorems}

\textbf{Assumption 1.} For any specific $\theta_t$, the maximum discrepancy between any individual gradient vector and the average gradient vector remains bounded by $r$ ensuring that the fairness promoting objective is $L_H$\ smooth. The bound 
$r$ effectively quantifies the allowable deviation, linking directly to $L_H$\, which controls the gradient’s variation and thus contributes to stable and fair gradient updates across all participating nodes. Consequently, 
\begin{equation}   
\max_{1 \leq i \leq M} \left\| \nabla H_i(\theta_t) - \frac{1}{M} \sum_{k=1}^M \nabla H_k(\theta_t) \right\| \leq r\,.
\end{equation}
\\\textbf{Nonconvex functions.}
We first examine the scenario where the objective function \( H(\theta) \) exhibits nonconvex properties with respect to \( \theta \). Under Assumption 1, we formulate Theorem~\ref{thm:theo,1}, which asserts that proposed algorithm will reliably converge to a stationary point of \( H(\theta) \).

\begin{theorem}
\label{thm:theo,1}
Let \( M \) be the total number of gradients, $1-p$ be the screening ratio, i.e., in total \( (1-p)m \) gradients will be discarded, and Gaussian noise is added with variance \( \sigma^2 \) for DP to each gradient, then the expected squared norm of the gradient satisfies:
\begin{equation}
\frac{1}{T} \sum_{t=0}^{T-1} \|\nabla H(\theta_t)\|^2 \leq \frac{2L}{T} \left( H(\theta_0) - H(\theta^*) \right) + \frac{\sigma^2 d}{p M} \,.
\end{equation}
\end{theorem} 

\noindent\textbf{Assumption 2.} $H(\theta)$ is convex. 

The convex assumption guarantees that any local minimum is a global minimum. 
%
Theorem~\ref{thm:theo,2} provides a convergence guarantee for our algorithm when the objective function \( H \) is convex.
\begin{theorem}
\label{thm:theo,2}
If the fraction of Byzantine nodes is \( \alpha < \frac{1}{3} \), then the expected value of the objective function at the output, \( \theta_T \), is bounded by:
\begin{equation}
\mathbb{E}[H(\theta_T)] - H(\theta^*) \leq \frac{B}{{T}} + \frac{\sigma^2 d}{p M}, 
\end{equation}
where \( \theta^* \) is the global minimizer of \( H \), and \( B \) is a constant that depends on the initial distance to the optimum and the Lipschitz constant \( L \).
\end{theorem}

\noindent
\textbf{Remark 1.}
The above analysis provides robust convergence guarantees in both non-convex and convex scenarios. Theorem~\ref{thm:theo,1} ensures convergence to a stationary point under non-convexity, accounting for DP and TNBS. Theorem~\ref{thm:theo,2} extends this to convex settings, showing convergence at a rate of \( \frac{1}{\sqrt{T}} \), highlighting the algorithm's efficiency in convex optimization tasks.


\section{Experiments}

This section compares the proposed method with several benchmarks, such as trimmed mean (TM) \cite{yin2018byzantine}, Krum \cite{blanchard2017machine}, coordinate-wise trimmed mean (CWTM) \cite{yin2018byzantine}, and a recently proposed H-nobs framework \cite{zhou2024h}. For fair comparison, all benchmark methods also utilize DP encryption and optimizes the fairness-promoting objective function.
\subsection{Experimental setup}

\begin{figure}[htb]
\centering
\begin{subfigure}{0.233\textwidth}
\includegraphics[width=\linewidth]{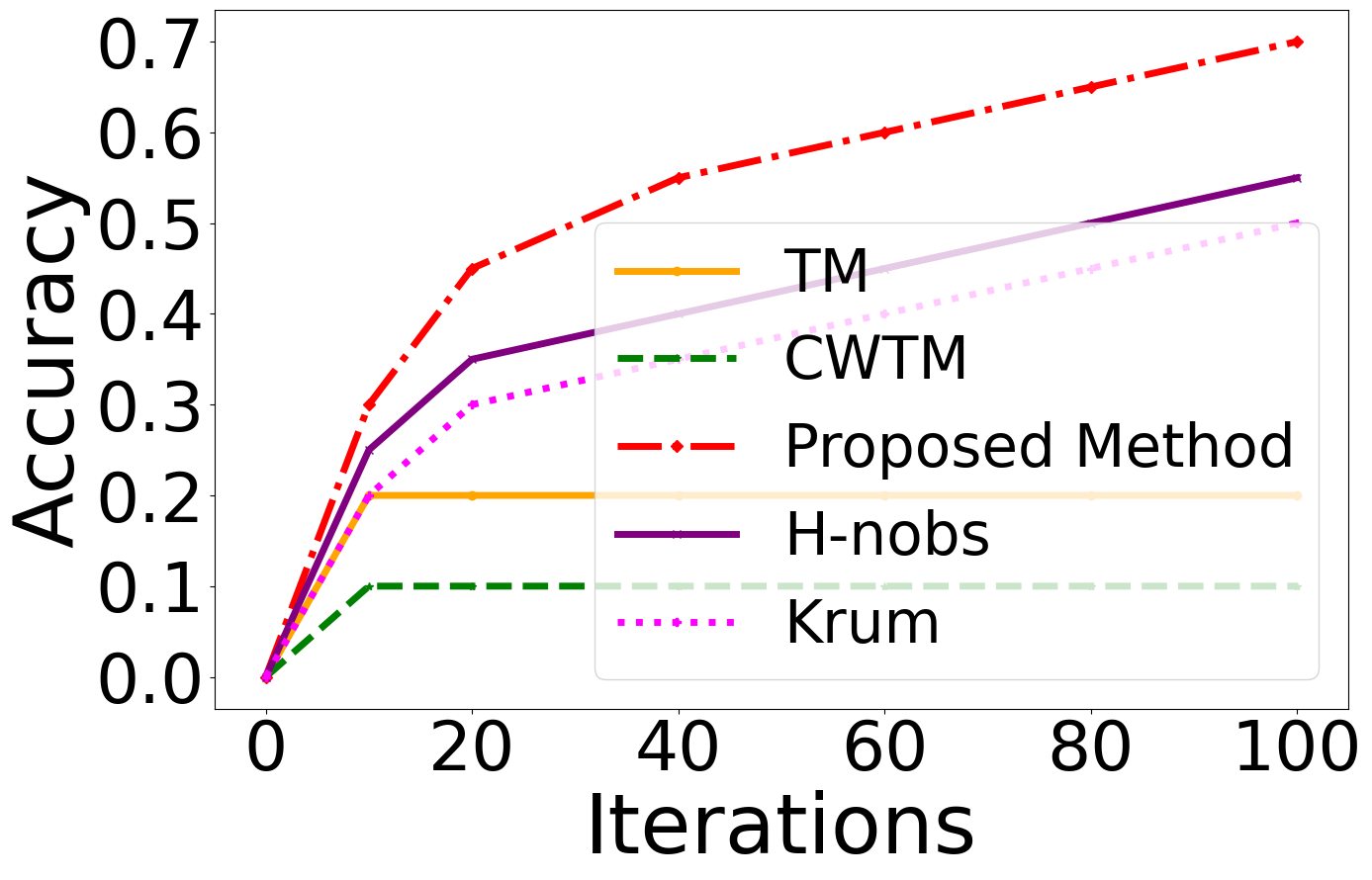}
\caption{No Attack}
\label{fig:noattM}
\end{subfigure}
\begin{subfigure}{0.233\textwidth}
\includegraphics[width=\linewidth]{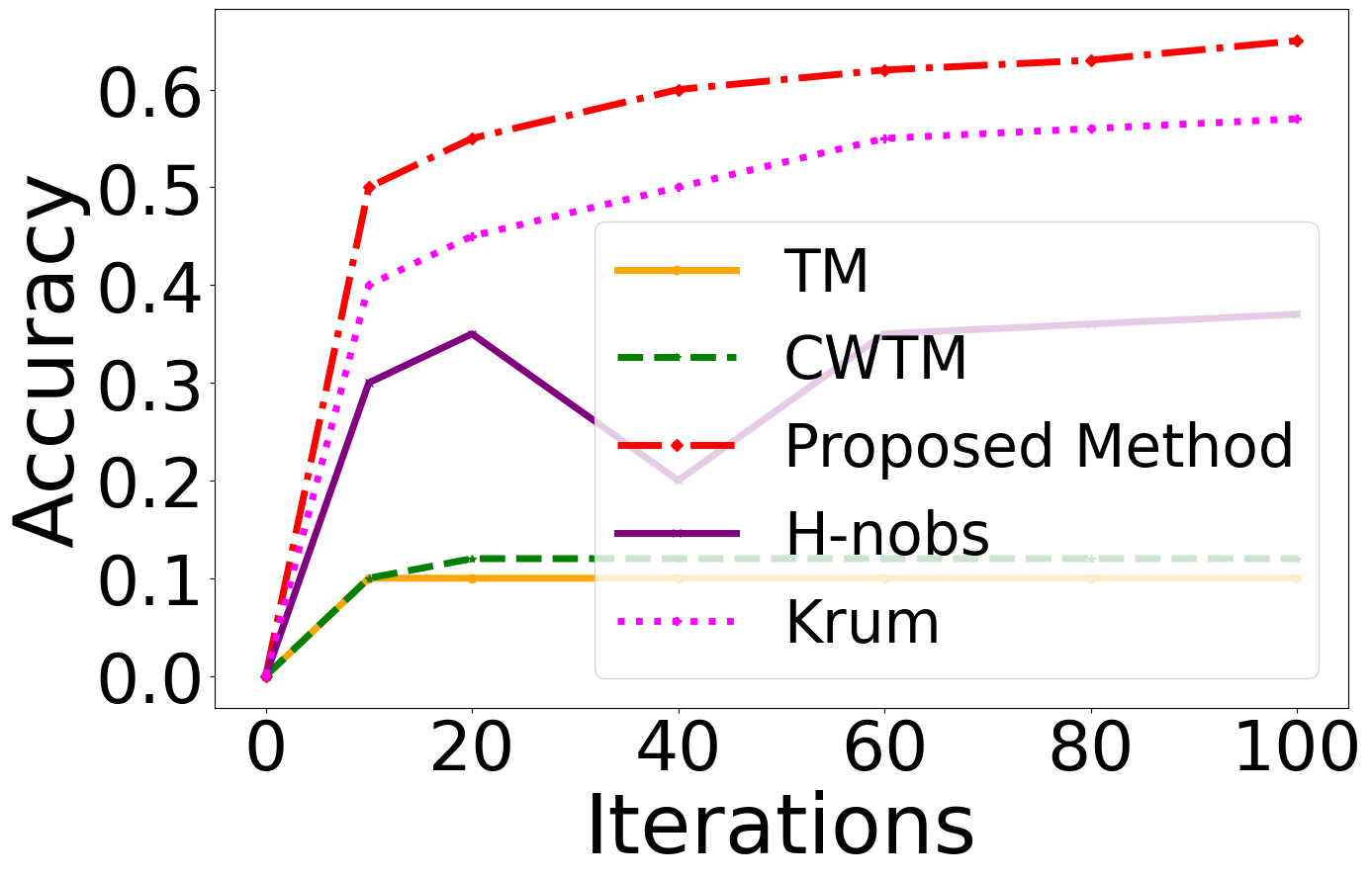}
\caption{Sign Flipping Attack}
\label{fig:SFattM}
\end{subfigure}


\begin{subfigure}{0.233\textwidth}
\includegraphics[width=\linewidth]{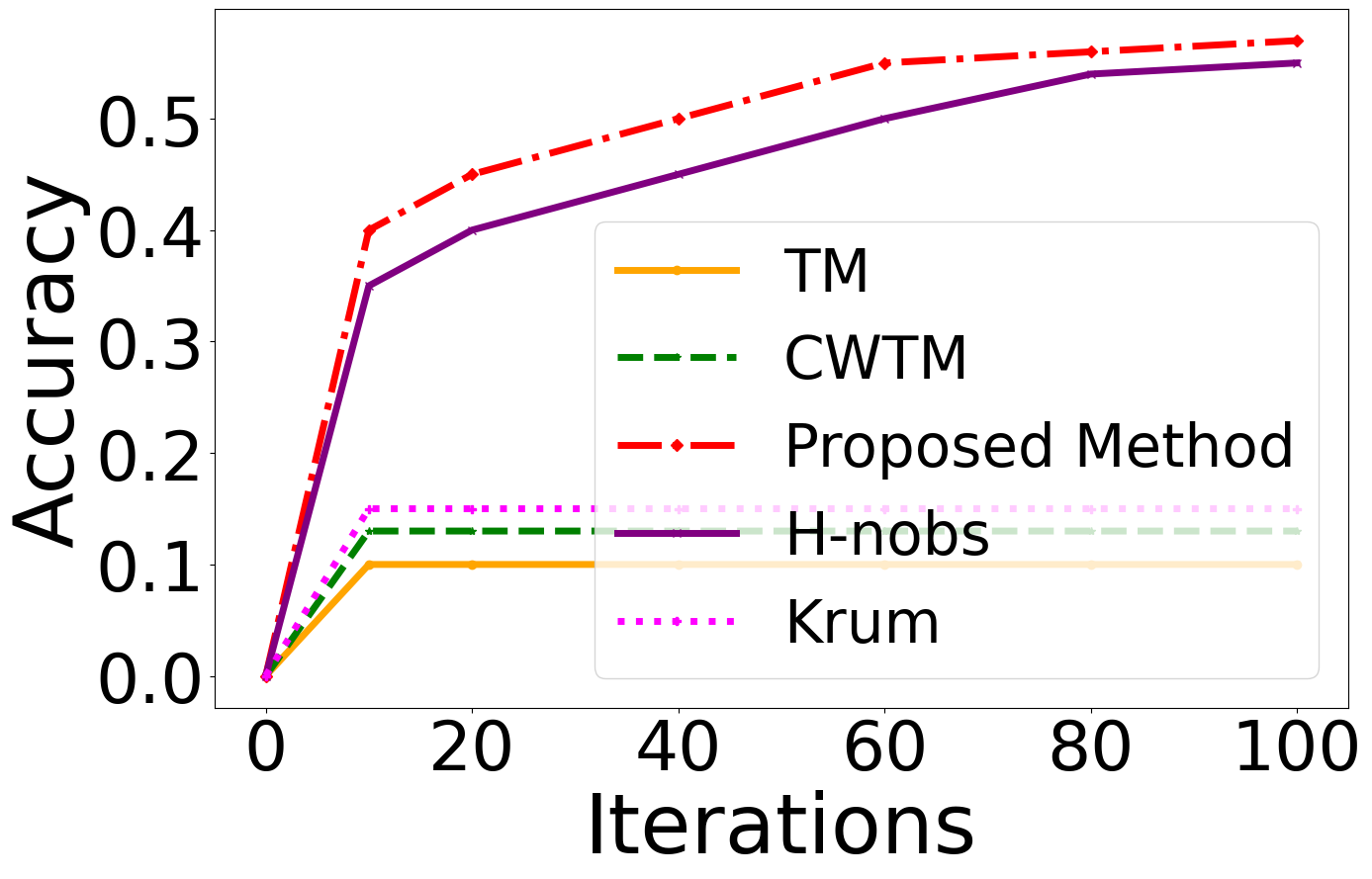}
\caption{Label Flipping Attack}
\label{fig:LFattM}
\end{subfigure}
\begin{subfigure}{0.233\textwidth}
\includegraphics[width=\linewidth]{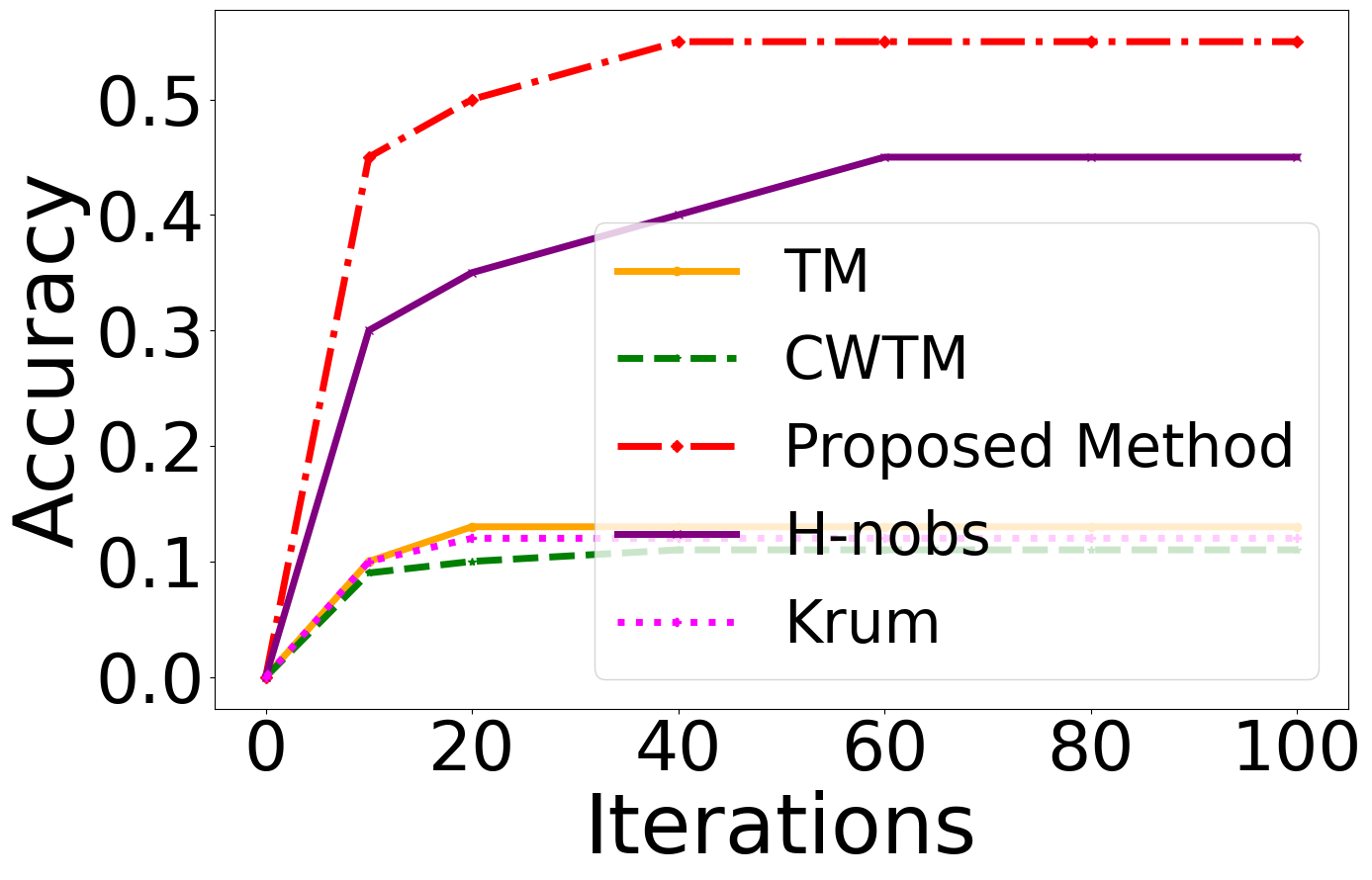}
\caption{Gaussian Attack}
\label{fig:GausattM}
\end{subfigure}

\caption{Comparison of model accuracy of proposed method with the other benchmarks for the MNIST dataset.}
\label{fig:comparisonMnist}
\end{figure}

Two datasets, MNIST \cite{726791} and Spam \cite{misc_spambase_94} are adopted in the experiments. The Spam dataset contains $4,601$ email text messages, labeled as either spam (1) or not spam (0). The MNIST dataset, consisting of 70,000 grayscale images of handwritten digits from 0 to 9. For each dataset, we allocate two-thirds for training and the remaining one-third for testing. In the experiments, we consider a total of 20 clients, among which 4 are Byzantine. The data distributions among agents are non-iid. For the Spam dataset \ref{fig:comparison}, data is evenly split among the clients with 4 nodes exclusively contain spam emails (labelled as 1) and the remaining 16 nodes contain non-spam emails (labelled as 0). For the MNIST dataset \ref{fig:comparisonMnist}, each digit is represented equally across exactly two clients. We keep the parameters $\eta=1$, $q=1$ and $\sigma^2=0.2$ throughout the experiments. 


\subsection{Results}

\begin{figure}[htb]
\centering
\begin{subfigure}{0.23\textwidth}
\includegraphics[width=\linewidth]{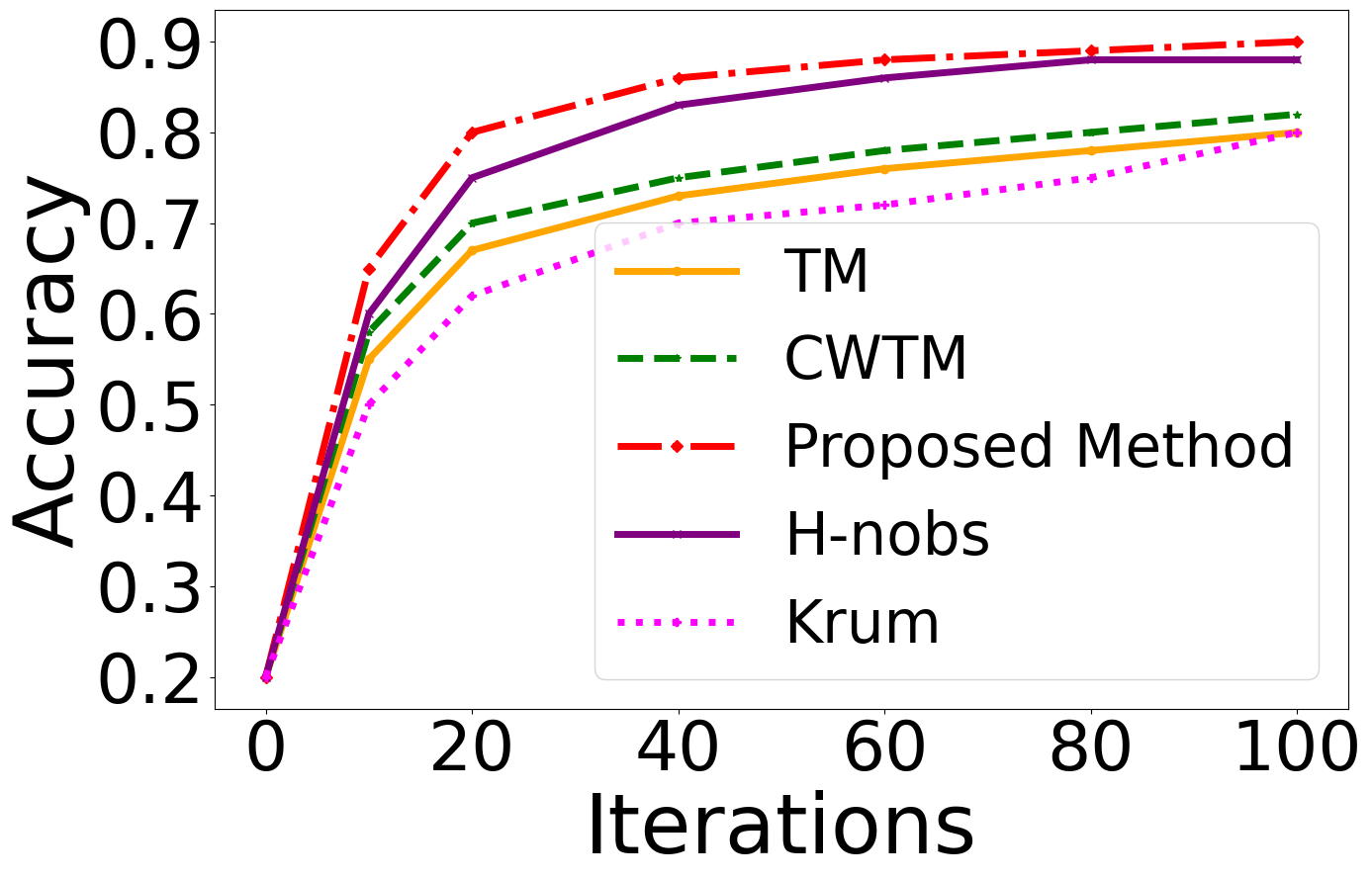}
\caption{No Attack}
\label{fig:noatt}
\end{subfigure}
\begin{subfigure}{0.23\textwidth}
\includegraphics[width=\linewidth]{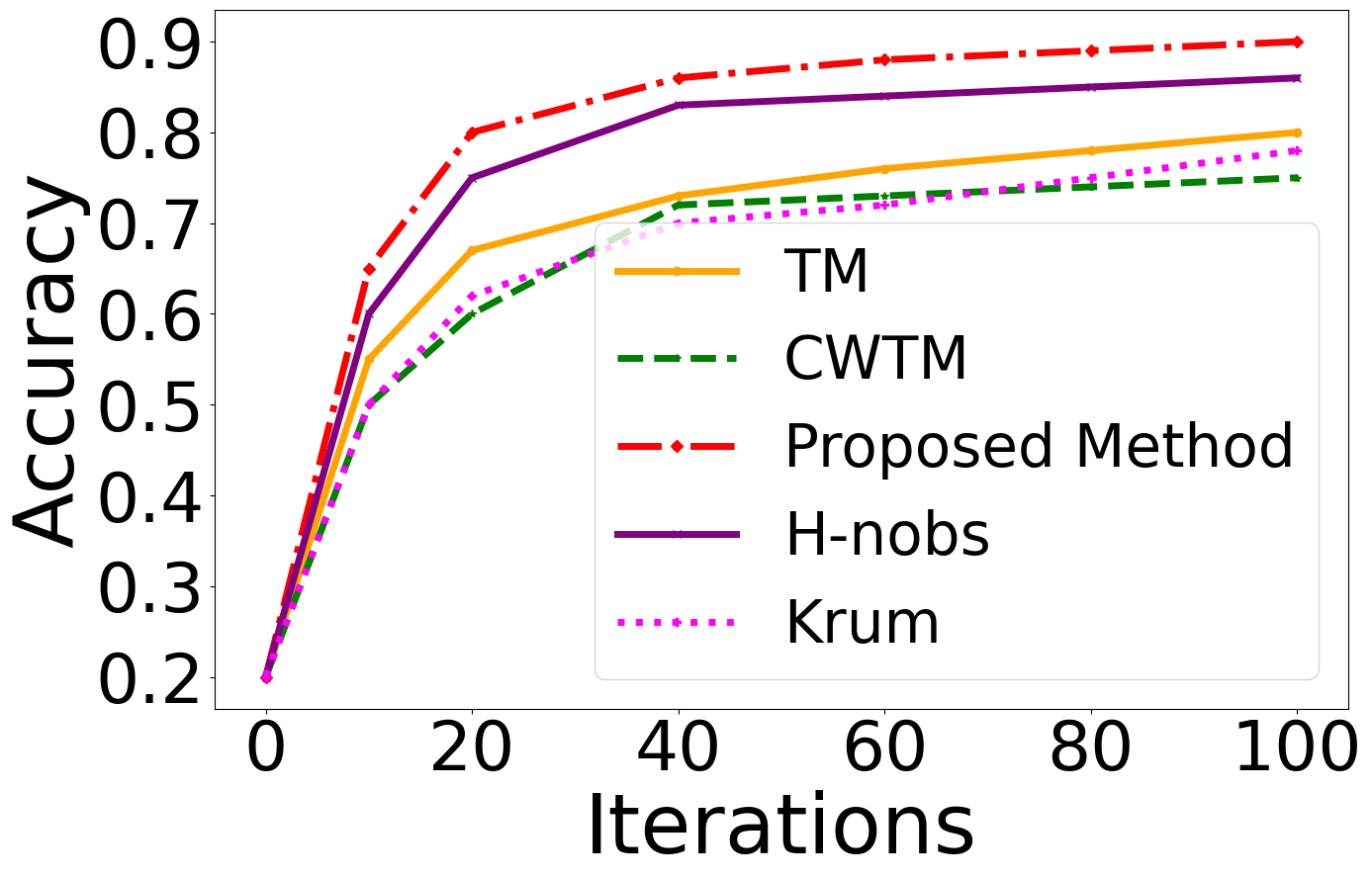}
\caption{Sign Flipping Attack}
\label{fig:SFatt}
\end{subfigure}
\begin{subfigure}{0.23\textwidth}
\includegraphics[width=\linewidth]{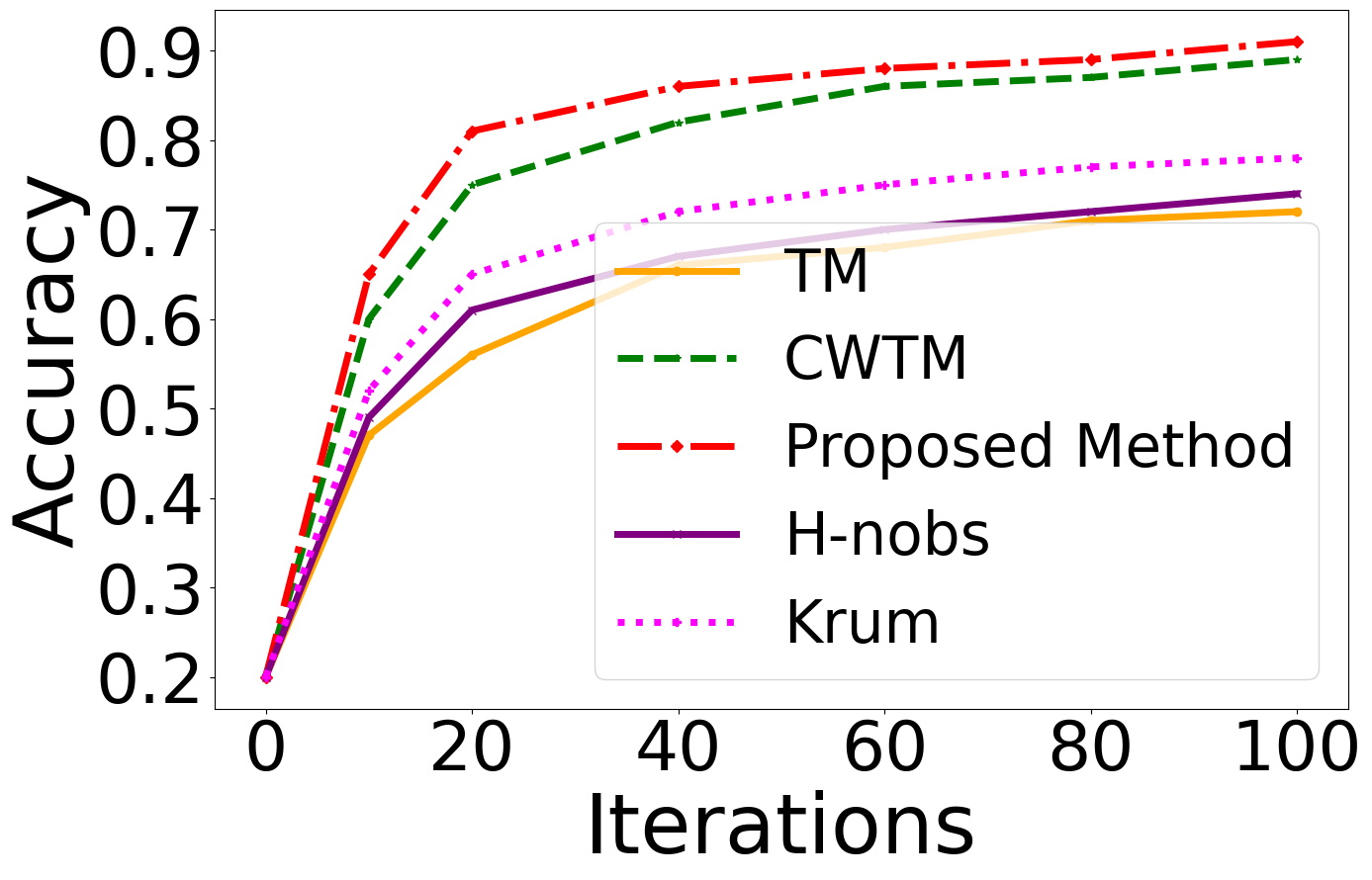}
\caption{Label Flipping Attack}
\label{fig:LFatt}
\end{subfigure}
\begin{subfigure}{0.23\textwidth}
\includegraphics[width=\linewidth]{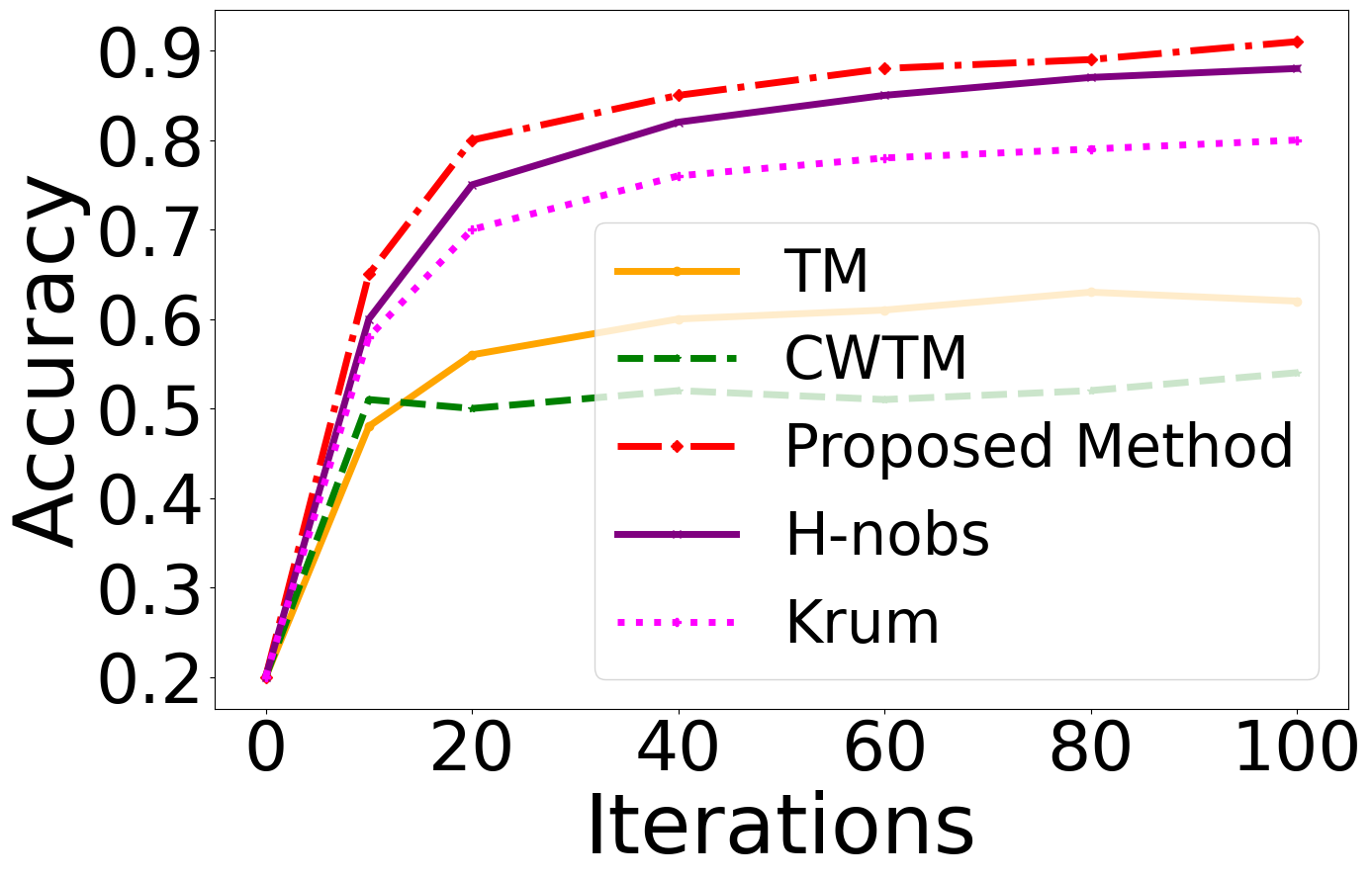}
\caption{Gaussian Attack}
\label{fig:Gausatt}
\end{subfigure}
\caption{Comparison of model accuracy of proposed method with the other benchmarks for the Spam based dataset.}
\label{fig:comparison}
\end{figure}

Compared with benchmark methods, the proposed method consistently outperforms the other benchmarks in all tests. Taking the Spam dataset as an example, under normal conditions without attacks (Fig.~\ref{fig:noatt}), 
the proposed method performs similarly to the others, showing high accuracy and stability. During the sign flipping attack (Fig.~\ref{fig:SFatt}) and the label flipping attack (Fig.~\ref{fig:LFatt}), 
it shows best resilience against attack, maintaining higher accuracy and stability quickly. For the Gaussian attack with an attack scale of $100$ (Fig.~\ref{fig:Gausatt}), our method still reaches the highest accuracy. These results demonstrate the robustness of our method, improving both privacy and fairness in adversarial settings. 

\begin{table}[ht]
\centering
\setlength{\tabcolsep}{1.8pt}
\renewcommand{\arraystretch}{1.7}
\caption{Accuracy (variance) performance of different methods with different $q$ values.}
\label{fairness}
\scalebox{.9}{
\begin{tabular}{rrrrrr}
\toprule
\bf $q$ parameter & \multicolumn{1}{c}{\bf TNBS}  & \multicolumn{1}{c}{\bf NBS} & \multicolumn{1}{c}{\bf Krum} & \multicolumn{1}{c}{\bf TM} & \multicolumn{1}{c}{\bf CWTM} \\
\midrule
$q = 0$  &\textbf{ 92.6(189)} & 92.5(396) & 90.1(320) & 89.0(359) &89.0(359)\\ 
$q = 0.5$ & \textbf{92.4(156)} & 92.3(368) & 89.2(200) & 84.1(276)& 89.0(359) \\ 
$q = 1$   & \textbf{92.0(116)} & 89.1(350) & 75.0(63) & 59.0(240) & 50.0(292) \\ 
\bottomrule

\end{tabular}
}
\end{table}


We evaluate fairness in Table~\ref{fairness} under Gaussian attack by examining the variance in accuracy performance across different clients. A lower variance in accuracy among clients indicates a more fair distribution of the trained model. On the overall accuracy, the proposed method is always better than TM and CWTM. Compared with NBS-based algorithm, only when $q=0.5$, we have a negligible performance degradation.

\begin{figure}[ht]
  \centering
  \includegraphics[width=0.32\textwidth]{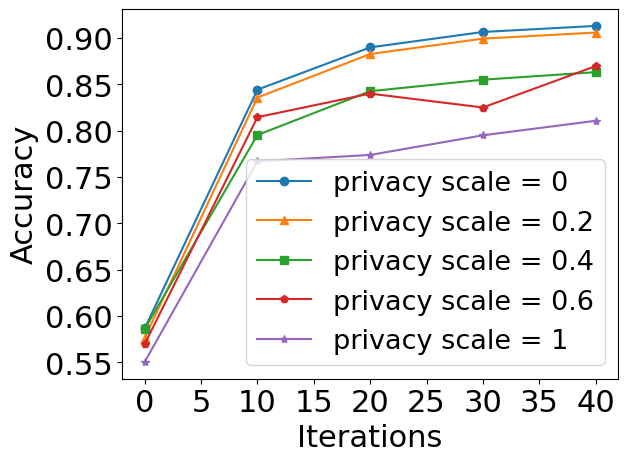}
  \caption{Trade-off between privacy and accuracy.}
  \label{fig:Privacy}
\end{figure}

The impact of DP on model accuracy is evaluated using the Spam dataset. As shown in Figure \ref{fig:Privacy}, there is a accuracy-privacy trade-off. Models with no added noise achieves the highest accuracy and increasing noise levels reduces accuracy. However, even at the highest noise level, the model retains significant learning capability.


\section{Conclusion}

This paper addresses three key challenges in federated learning: client-level fairness, privacy, and Byzantine attack resilience. We achieve fairness through $q$-fair federated learning, ensure privacy with Gaussian noise-added differential privacy, and defend against attacks using a two-sided norm-based screening method. Our approach is both theoretically grounded and empirically validated across various attack scenarios.


\bibliography{references}


\appendix

\section{Appendix}\label{pend_A}
\subsection{Proof of Lemma 1}

Let $M$ be the indices of honest gradients, and $B$ as the indices of Byzantine gradients, with $|B| = \alpha m$. The screening removes the highest $1-\frac {1-p}{2} m$ and lowest $\frac{1-p}{2} m$ gradients by norm, resulting in the inclusion of the middle $p$ proportion of gradients.

After the screening, the remaining gradients are aggregated to form $G$, that is,
\begin{equation}
    \label{e14}
    G = \frac{1}{m(1-\alpha)} \sum_{i \in U} g_i \,,
\end{equation}
where $U$ is the set of indices of gradients after screening.
%
Then we express $G$ in terms of the deviation from $\nabla H$ and get
   \begin{equation}
   \label{e15}
    G - \nabla H = \frac{1}{m(1-\alpha)} \left(\sum_{i \in M \cap U} g_i + \sum_{i \in B \cap U} g_i \right) - \nabla H \,.
    \end{equation}
By rearranging \eqref{e15} to focus on deviations, we can obtain that  
\begin{equation}
\begin{split}
    G - \nabla H &= \frac{1}{m(1-\alpha)} 
    \Bigg(\sum_{i \in M \cap U} (g_i - \nabla H) \\
    &\quad + \sum_{i \in B \cap U} (g_i - \nabla H) \Bigg).
\end{split}
\end{equation}

For honest gradients $i \in M \cap U$, their deviation from $\nabla H$ is bounded by the gradient norms. Assuming the worst case manipulation for Byzantine gradients $i \in B \cap U$, we have
    \begin{equation}
    \sum_{i \in B \cap U} (g_i - \nabla H) \leq \max_{i \in M} \|g_i - \nabla H\|
    \,.
    \end{equation}
Combining the honest and Byzantine components to derive the inequality, we can obtain that
    \begin{equation}
    \|G - \nabla H\| \leq \frac{2\alpha}{1-p} \|\nabla H\| + \max_{i \in M} \|g_i - \nabla H\| + \sigma
    \,.
    \end{equation}

In this proof, we have rigorously demonstrated the effectiveness of the proposed mechanism under the conditions stipulated for FL involving Byzantine attacks, fairness and differential privacy. Specifically, we have shown that the aggregated gradient \( G \), as derived through TNBS, remains within a controlled deviation from the true gradient \( \nabla H \). This bounded deviation quantitatively ensures the resilience of the learning process against adversarial influences, while simultaneously maintaining the privacy and fairness across the network. Hence, our findings validate the robustness of the proposed framework, confirming its theoretical and practical viability for deployment in environments susceptible to Byzantine behaviors and privacy concerns.


\subsection{Proof of Theorem~\ref{thm:theo,1}}
\label{pend_B}
Let $m$ be the total number of gradients, and $p = (1 - 2\beta)m$ be the number of gradients after excluding the highest and lowest $\beta m$ gradients. Given that Gaussian noise with variance $\sigma^2$ is added for DP to each gradient, the expected squared norm of the gradient satisfies
\begin{equation}
\begin{split}
    E\left[\frac{1}{T} \sum_{t=0}^{T-1} \|\nabla H(\theta_t)\|^2\right] 
    &\leq \frac{2L}{T} (H(\theta_0) - H(\theta^*)) \\
    &\quad + \frac{\sigma^2 d}{(1-2\beta)m}.
\end{split}
\end{equation}

The gradient descent update under the influence of both Byzantine attacks and added noise can be written as
\begin{equation}
    \theta_{t+1} = \theta_t - \eta \hat{G}_t
    \,,
\end{equation}
where $\hat{G}_t$ is the aggregated gradient post Byzantine filtering and noise addition.
The expected gradient, considering the Byzantine robust mechanism and noise, can be approximated by
\begin{equation}
    E[\hat{G}_t] \approx \nabla H(\theta_t) - \text{Bias}_{\text{Byz}} + \text{Noise}_{\text{DP}}
    \,,
\end{equation}
where $\text{Bias}_{\text{Byz}}$ represents the error introduced by not filtering out all Byzantine gradients perfectly, and $\text{Noise}_{\text{DP}}$ represents the zero-mean Gaussian noise.

By decomposing the squared norm of the expected gradient into signal and noise components, we have
\begin{equation}
    \label{decomp}
    E[\|\hat{G}_t\|^2] = \|\nabla H(\theta_t)\|^2 + \|\text{Bias}_{\text{Byz}}\|^2 + \sigma^2,
\end{equation}
assuming independence between the honest gradient, the bias, and the noise.

Using a Lyapunov function $V(t) = H(\theta_t) - H(\theta^*)$ to evaluate the convergence, we can obtain that
\begin{equation}
    \label{lyapunov}
    E[V(t+1) - V(t)] \leq -\eta \nabla H(\theta_t)^\top \hat{G}_t + \frac{L}{2} \eta^2 E[\|\hat{G}_t\|^2]
    \,.
\end{equation}

Plugging \eqref{lyapunov} in the \eqref{decomp}, we get

\begin{equation}
\begin{split}
    E[V(t+1) - V(t)] &\leq -\eta \|\nabla H(\theta_t)\|^2 \\
    &\quad + \eta^2 L \big(\|\nabla H(\theta_t)\|^2  
    + \|\text{Bias}_{\text{Byz}}\|^2 \\
    &\quad + \sigma^2\big).
\end{split}
\end{equation}

Summing the drift over $T$ iterations and rearranging to isolate the gradient norm term, it becomes
\begin{equation}
    \sum_{t=0}^{T-1} \|\nabla H(\theta_t)\|^2 \leq \frac{H(\theta_0) - H(\theta^*)}{\eta(1-\eta L)} + T \eta L (\|\text{Bias}_{\text{Byz}}\|^2 + \sigma^2)
    \,.
\end{equation}

For sufficiently small $\eta$, the convergence rate to a stationary point is influenced by the level of noise and the effectiveness of the Byzantine robust mechanism, specifically,
\begin{align}
E\left[\frac{1}{T} \sum_{t=0}^{T-1} \|\nabla H(\theta_t)\|^2\right] 
&\leq \frac{2L}{T} (H(\theta_0) - H(\theta^*)) \notag \\
&\quad + \frac{\sigma^2 d}{(1-2\beta)m} \,.
\end{align}

Therefore, Theorem~\ref{thm:theo,1} validates that the proposed framework achieves a predictable rate of convergence to a stationary point, incorporating the robustness against Byzantine influences, fairness and the differential privacy. This proof underscores the framework's capability to handle adversarial disruptions effectively while advancing towards optimal model parameters in non-convex settings.


\subsection{Proof of Theorem~\ref{thm:theo,2}}
\label{pend_C}

Given the convexity of the global objective function \( H(\theta) \), the fraction of Byzantine nodes being less than \(\frac{1}{3}\), and the DP mechanisms in place, the expected value of the objective function at the output, \( \theta_T \), after \( T \) iterations, is bounded by:
\begin{equation}
E[H(\theta_T)] - H(\theta^*) \leq \frac{B}{T} + \frac{\sigma^2 d}{(1 - 2\beta)m} \,,
\end{equation}
where \( \theta^* \) is the global minimizer of \( H \), \( B \) is a constant dependent on the initial distance to the optimum, and the Lipschitz constant \( L \).

Let $\hat{G}_t$ be the aggregated gradient post Byzantine filtering and differential privacy noise addition. 
Since \( H \) is convex, for any \( \theta_t \) and \( \theta^* \), we have
\begin{equation}
    H(\theta^*) \geq H(\theta_t) + \nabla H(\theta_t)^\top (\theta^* - \theta_t).
\end{equation}

Given the updating rule
\begin{equation}
    \theta_{t+1} = \theta_t - \eta_t \hat{G}_t \,,
\end{equation}
combining this with the convexity of \( H \) and rearranging terms, we get
\begin{equation}
    H(\theta_{t+1}) \leq H(\theta_t) - \eta_t \nabla H(\theta_t)^\top \hat{G}_t + \frac{L}{2} \eta_t^2 \|\hat{G}_t\|^2.
\end{equation}

Taking expectations and using the boundedness of the gradients by \( S \) and the variance introduced by differential privacy:
    \begin{equation}
    E[H(\theta_{t+1})] \leq H(\theta_t) - \eta_t \|\nabla H(\theta_t)\|^2 + \frac{L}{2} \eta_t^2 (S^2 + \sigma^2)
    \,.
  \end{equation}

Summing over \( t \) from \( 0 \) to \( T-1 \) and averaging, we derive
\begin{equation}
    E[H(\theta_T)] - H(\theta^*) \leq \frac{H(\theta_0) - H(\theta^*)}{T \eta_t} + \frac{L \eta_t (S^2 + \sigma^2)}{2}.
\end{equation}

Note that optimal step sizes and assumptions on the decay of \( \eta_t \) can minimize the right-hand side of the inequality, ensuring convergence to \( \theta^* \).

In conclusion, Theorem~\ref{thm:theo,2} demonstrates that our framework ensures robust and efficient convergence in convex environments, even in the presence of Byzantine attacks and while maintaining strict privacy standards. This proof conclusively establishes that our algorithm not only meets theoretical expectations for convergence but also provides practical resilience, fairness and privacy, crucial for real-world applications.


\end{document}